# AN APPROACH TO SPEED-UP THE WORD SENSE DISAMBIGUATION PROCEDURE THROUGH SENSE FILTERING

Alok Ranjan Pal,[1] Anupam Munshi[1] and Diganta Saha[2]

[1]Dept. of Computer Science and Engineering College of Engineering and Management, Kolaghat,West Bengal, India.

[2]Dept. of Computer Science and Engineering , Jadavpur University,Kolkata, India

## Abstract

*In this paper, we are going to focus on speed up of the Word Sense Disambiguation procedure by filtering the relevant senses of an ambiguous word through Part-of-Speech Tagging. First, this proposed approach performs the Part-of-Speech Tagging operation before the disambiguation procedure using Bigram approximation. As a result, the exact Part-of-Speech of the ambiguous word at a particular text instance is derived. In the next stage, only those dictionary definitions (glosses) are retrieved from an online dictionary, which are associated with that particular Part-of-Speech to disambiguate the exact sense of the ambiguous word.*

*In the training phase, we have used Brown Corpus for Part-of-Speech Tagging and WordNet as an online dictionary. The proposed approach reduces the execution time upto half (approximately) of the normal execution time for a text, containing around 200 sentences. Not only that, we have found several instances, where the correct sense of an ambiguous word is found for using the Part-of-Speech Tagging before the Disambiguation procedure.*

## Key words

## 1. INTRODUCTION

In human languages all over the world, there are a lot of words having different meanings depending on the contexts. Word Sense Disambiguation (WSD) [1-8] is the process for identification of actual meaning of an ambiguous word based on distinct situations. As for example, the word "Bank" has several meanings, such as "place for monitory transaction", "reservoir", "turning point of a river", and so on. Such words with multiple meanings are ambiguous in nature. The process to decide the appropriate meaning of an ambiguous word for a particular context is known as Word Sense Disambiguation. People have inborn ability to sense the actual meaning of an ambiguous word in a particular context. But the machines do this job by some pre-defined rules or statistical methods.

Two types of learning procedures are commonly used for Word Sense Disambiguation procedure. First, Supervised Learning, where a learning set is considered for the system to predict the actual meaning of an ambiguous word using a few sentences, having a specific meaning for that particular word. A system finds the actual meaning of an ambiguous word for a particular context based on that defined learning set. In this method, learning set is created manually. As a result, it is unable to generate fixed rules for all the systems. Therefore, the actual meaning of an ambiguous word in a given context can't be detected always. Supervised learning derives partially correct result, if the learning set does not contain sufficient information for all possible senses of the ambiguous word. Even, it fails to show the result, if there is no information in the predefined database.

In Unsupervised learning, an online dictionary is taken as learning set avoiding the inefficiency of Supervised learning. "WordNet"[9-15] is the most widely used online dictionary maintaining "words and related meanings" as well as "relations among different words".

But in Unsupervised Learning procedures, commonly used for Sense Disambiguation, all the dictionary definitions (glosses) of the ambiguous word are considered from the Dictionary. These glosses are of different types of Part-of-Speech (POS), such as noun, verb, adjective and adverb. In case of commonly used Unsupervised Learning procedures like Lesk Algorithm [16, 17], all the glosses of different Part-of-Speech are considered, which takes some unnecessary additional execution time. As an ambiguous word carries a specific Part-of-Speech in a particular context, we have gone through Part-of-Speech Tagging [18-31] before the WSD procedure. As a result, only the glosses of the related Part-of-Speech are considered.

Using this approach, we have observed two types of betterment in the output. First, the execution of the disambiguation procedure becomes faster and second, as the relevant glosses are filtered, accuracy in disambiguated sense is increased.

Organization of rest of the paper is as follows: Section 2 is about the Theoretical Background of the proposed approach; Section 3 describes the Implementation Background; Section 4 describes the Proposed Approach in detail; Section 5 depicts the experimental results along with comparison; Section 6 represents Conclusion of the paper.

## 2. THEORETICAL BACKGROUND

The most common Unsupervised WSD algorithm is Lesk Algorithm, which uses WordNet as an online dictionary. The algorithm is described below in brief:

### 2.1 Preliminaries of Lesk Algorithm

Typical Lesk Algorithm selects a short phrase from the sentence containing an ambiguous word. Then, dictionary definition (gloss) of each of the senses of the ambiguous word is compared with glosses of the other words in that particular phrase. An ambiguous word is being assigned with the particular sense, whose gloss has the highest number of overlaps (number of common words) with the glosses of the other words of the phrase.

**Example 1:** "Ram and Sita everyday go to bank for withdrawal of money."
Here, the phrase is taken depending on window size (number of consecutive words). If window size is 3, then the phrase would be "go bank withdrawal". All other words are being discarded as "stop words".

Consider the glosses of all words presented in that particular phrase are as follows:
Suppose, the number of senses of "Bank" is 2 such as 'X' and 'Y' (refer Table 1).
The number of senses of "Go" is 2 such as 'A' and 'B' (refer Table 2).
And the number of senses of "Withdrawal" is 2 such as 'M' and 'N' (refer Table 3).

| Keyword | Probable sense |
|---------|----------------|
| Bank    | X              |
|         | Y              |

Table 1.  Probable Senses of "Bank".

| Word | Probable sense |
|------|----------------|
| Go   | A              |
|      | B              |

Table 2.  Probable Senses of "Go".

| Word       | Probable sense |
|------------|----------------|
| Withdrawal | M              |
|            | N              |

Table 3.  Probable Senses of "Withdrawal".

Consider the word "Bank" as a keyword. Number of common words is measured in between a pair of sentences.

| Pair of Sentences | Common number of Words |
|---|---|
| X and A | A' |
| X and B | B' |
| Y and A | A'' |
| Y and B | B'' |
| X and M | M' |
| X and N | N' |
| Y and M | M'' |
| Y and N | N'' |

Table 4.  Comparison chart between pair of sentences and common number of words within a particular pair.

Table 4 shows all possibilities using sentences from Table 1, Table 2, Table 3, and number of words common in each possible pair.

Finally, two senses of the keyword "Bank" have their counter readings (refer Table 4) as follows:

$$X \text{ counter}, X_C = A' + B' + M' + N'.$$
$$Y \text{ counter}, Y_C = A'' + B'' + M'' + N''.$$

Therefore, higher counter value would be assigned as the sense of the keyword "Bank" in the particular sentence. This strategy believes that surrounding words have the same senses as of the keyword.

## 2.2    Simple (unsmoothed) N-gram and Bigram Model

N-gram [32] is used to compute the probability of a complete string of words (which can be represented either as $w_1 \ldots w_n$ or $w_1^n$). If each word, occurring in its correct location, is considered as an independent event, the probability is represented as:
P ($W_1$, $W_2$,……,$W_{n-1}$, $W_n$).

The chain rule to decompose the probability would be:

$P(W_1^n)=P(W_1)P(W_2|W_1)P(W_3|W_1^2)\ldots P(W_n|W_1^{n-1})$.

But, computing the probability like $P(W_n|W_1^{n-1})$ is not easy for a long sequence of preceding words. The Bigram model approximates the probability of a word with respect to all the previous words $P(W_n|W_1^{n-1})$ by the conditional probability of the just preceding word $P(W_n | W_{n-1})$.

For example, instead of computing the probability P(rabbit | Just the other day I saw a), the probability is approximated by P(rabbit | a).

## 3. IMPLEMENTATION BACKGROUND

This paper adopts the basic ideas from typical Lesk algorithm by introducing some modifications.

### 3.1 Simplified Lesk Approach

In this approach, the glosses of only the keyword are considered for a specific sentence instead of all words. Number of common words is being calculated between the specific sentence and each dictionary definition of the particular keyword.

- Consider, earlier mentioned sentence of "Example 1" as follows: "Ram and Sita everyday go to bank for withdrawal of money."
- The instance sentence would be "Ram Sita everyday go bank withdrawal money" after discarding the "stop words" like "to", "for" and so on.
- If "Bank" is considered as the keyword and its two senses are X and Y (refer Table 1). Then, number of common words should be calculated between the instance sentence and each probable senses of "Bank" (refer Table 1).
- Number of common words found would be assigned to the counter of that sense of "Bank". Consider X-counter has the value I' and Y-counter has the value I".
- Finally, the higher counter value would be assigned as the sense of the keyword for the particular instance sentence.
- The dictionary definition (gloss) of the keyword would be taken from "WordNet".
- This approach also believes that entire sentence represents a particular sense of the keyword.

## 4. PROPOSED APPROACH

The proposed approach derives the actual sense of an ambiguous word in two steps. First, the input text is passed through the POS Tagging module, where the POS of the ambiguous word is derived. Second, the input sentence, containing the ambiguous word with derived POS is passed to WSD module, where the disambiguation operation is performed using Simplified Lesk Algorithm.

As the POS of the ambiguous word is derived before WSD operation, the selected dictionary definitions (glosses) are filtered from all the instances present in WordNet (as Noun, Verb, Adjective and Adverb instances). As a result, the disambiguation procedure becomes faster.

Not only that, as the POS of the ambiguous word is derived before the WSD operation, the disambiguation algorithm is applied on only the relevant glosses. As a result, the accuracy of the

disambiguation algorithm is increased. The detail explanation of the proposed approach is given below:

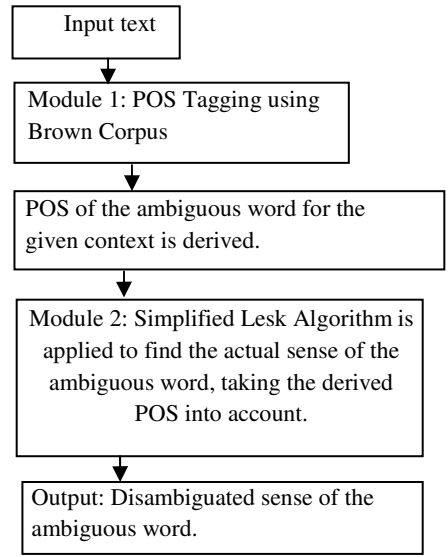

Figure 1. Modular representation of the overall approach

**Algorithm 1:** This algorithm (refer Figure 1) describes the overall approach. The first module is responsible for POS Tagging and the second module is responsible for WSD task.
Input: Input text, containing the ambiguous word.
Output: Disambiguated sense of the ambiguous word.

Step 1: Input text, containing the ambiguous word is passed to Module 1 for finding the POS of the ambiguous word.
Step 2: Simplified Lesk Algorithm is applied to find the actual sense of the ambiguous word, taking the derived POS into account.
Step 3: Stop.

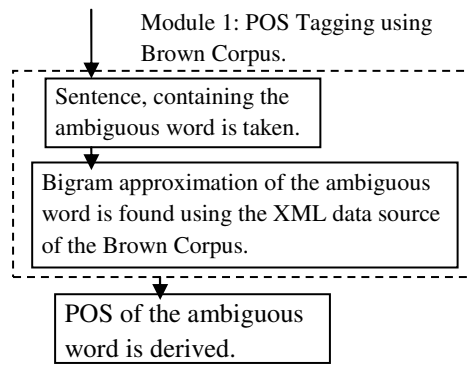

Figure 2. Implementation detail of Module 1 for POS Tagging

**Module 1: Algorithm 2:**

This algorithm (refer Figure 2) finds the POS of the ambiguous word using Brown Corpus. The maximum Time Complexity of the algorithm is $O(n^2)$, which is evaluated at step 2.
Input: Sentence, containing the ambiguous word.

Output: POS of the ambiguous word.

Step 1: Input sentence, containing the ambiguous word is taken.
Step 2: Bigram approximation of the ambiguous word is found using the XML data source of the Brown Corpus.
Step 3: POS of the ambiguous word is derived from the largest approximation value.
Step 4: Stop.

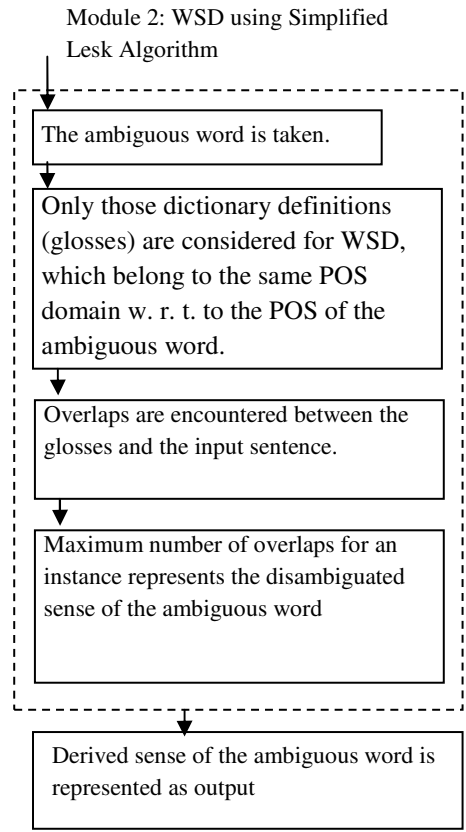

Figure 3.  Implementation detail of Module 2 for WSD procedure

**Module 2: Algorithm3:** This algorithm (refer Figure 3) derives the actual sense of an ambiguous word using the Simplified Lesk Algorithm. Time Complexity of the algorithm is $O(n^3)$, as finding the total number of overlaps between a particular gloss and the input sentence is of $O(n^2)$ complexity and this procedure is performed for all the n number of glosses.

Input: Ambiguous word with derived POS.

Output: Disambiguated sense of the ambiguous word.

Step 1: The ambiguous word is taken.
Step 2: Only those dictionary definitions (glosses) are considered from WordNet, which belong to the same POS domain w. r. t. to the POS of the ambiguous word.
Step 3: Overlaps are encountered between the glosses and the input sentence itself.
Step 4: The actual sense of the ambiguous word is derived from the maximum number of overlaps for an instance.
Step 5: Stop.

The proposed approach gives better result regarding the execution time and the accuracy of the result, which is described in the next section.

## 5.    OUTPUT AND DISCUSSION

The algorithm is tested on more than 100 texts of different lengths and categories. Average length of the texts is of 200 sentences and two ambiguous words are selected for testing, "Bank" and "Plant".

Next, the Simplified Lesk Algorithm is applied on the input text, containing the POS-tagged ambiguous word. As the POS of the ambiguous word is derived earlier, only those dictionary definitions are selected from WordNet for WSD process, which belong to the same POS domain w. r. t. the POS of the ambiguous word. As a result, the execution time of the WSD process becomes less (refer Table 5).

It is also observed that, as the relevant glosses are considered for the WSD process, accuracy of the disambiguated sense is increased (refer Text no. 10).
Some of the results for target word "Bank" are given in Table 5. All the sample texts are taken from "www.wikipedia.com".

| Text No. | Sentence in the text | D-sense with POS Tagging | D-sense without POS Tagging | E-time with POS Tagging (ms) | E-time without POS Tagging(ms) |
|---|---|---|---|---|---|
| 1 | 180 | Financial Institution | Financial Institution | 50 | 80 |
| 2 | 210 | Financial Institution | Financial Institution | 53 | 94 |
| 3 | 180 | Financial Institution | Financial Institution | 48 | 78 |
| 4 | 195 | Financial Institution | Financial Institution | 60 | 94 |
| 5 | 192 | Financial Institution | Financial Institution | 52 | 78 |
| 6 | 207 | Financial Institution | Financial Institution | 53 | 93 |
| 7 | 228 | Financial Institution | Financial Institution | 58 | 93 |
| 8 | 198 | Financial Institution | Financial Institution | 52 | 79 |
| 9 | 210 | Financial Institution | Financial Institution | 53 | 94 |

Table 5. Speed up analysis of WSD procedure for target word "Bank".

Note 1: D-sense means Disambiguated sense, E-time means Execution time, ms means Mille Second.

In the following sample test (Text no. 10), it is depicted that, in the given input text the ambiguous word "Plant" carries the actual sense (decided by human) as "Living Organism", which is in noun sense, but when the algorithm runs without POS Tagging, it derives the sense "Contact", which is in verb sense and obviously it is a wrong sense for this context according to human decision. This text is also taken from "www.wikipedia.com".

Accuracy measurement of WSD procedure using the proposed approach is described below with a sample text.

**Text no. 10:**

Plants, also called green plants, are living organisms of the kingdom Plantae including such multi cellular groups as flowering plants, conifers, ferns and mosses, as well as, depending on definition, the green algae, but not red or brown seaweeds like kelp, nor fungi or bacteria.
Green plants have cell walls with cellulose and characteristically obtain most of their energy from sunlight via photosynthesis using chlorophyll contained in chloroplasts, which gives them their green color. Some plants are parasitic and may not produce normal amounts of chlorophyll or photosynthesize. Plants are also characterized by sexual reproduction, modular and indeterminate growth, and an alternation of generations, although asexual reproduction is common, and some plants bloom only once while others bears only one bloom.

Precise numbers are difficult to determine, but as of 2010, there are thought to be 300–315 thousand species of plants, of which the great majority, some 260–290 thousand, are seed plants. Green plants provide most of the world's molecular oxygen and are the basis of most of the earth's ecologies, especially on land. Plants described as grains, fruits and vegetables form mankind's basic foodstuffs, and have been domesticated for millennia. Plants enrich our lives as flowers and ornaments. Until recently and in great variety they have served as the source of most of our medicines and drugs. Their scientific study is known as botany.

**Output:**

Target word: Plant.
Actual sense: Living Organism (Noun).
Derived sense with POS Tagging: Living Organism (Noun).
Derived sense without POS Tagging: Contact (Verb).

# 6. CONCLUSION AND FUTURE WORK

The proposed approach speeds up the WSD procedure by filtering the only relevant glosses and increases the accuracy of the WSD procedure as well. The execution time differences between the two cases (with and without POS Tagging procedure) might be increased, if few system calls and other system related tasks are handled properly. The obvious operations (loop, memory allocation, condition check, function call etc.) for POS Tagging took some time, which is included in the cited result. Otherwise, the actual time difference could have been better.

**Authors**

Alok Ranjan Pal has been working as an a Assistant Professor in Computer Science and Engineering Department of College of Engineering and Management, Kolaghat since 2006. He has completed his Bachelor's and Master's degree under WBUT. Now, he is working on Natural Language Processing

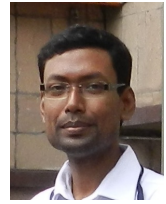

Mr. Anupam Munshi is a student of Information Technology Department of College of Engineering and Management, Kolaghat. His field of interest is AI, Soft Computing and NLP.

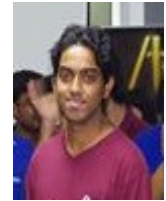

Dr. Diganta Saha is an Associate Professor in Department of Computer Science & Engineering, Jadavpur University. His field of specialization is Machine Translation/ Natural Language Processing/ Mobile Computing/ Pattern Classification.

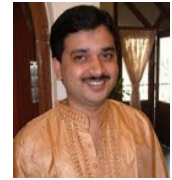